%% file: main.tex
\documentclass[letterpaper, 10 pt, journal, twoside]{IEEEtran}

\usepackage{cite}

\usepackage{hyperref}
\usepackage[T1]{fontenc}
\usepackage{breqn}
\usepackage{booktabs}
\usepackage{adjustbox}
\usepackage{multirow}
\usepackage{bbding}
\usepackage{colortbl}
\usepackage{amsmath,amssymb,amsfonts}
\usepackage{algorithmic}
\usepackage{graphicx}
\usepackage{xcolor}

\author{Yixiang Chen$^{1,2}$, Yan Huang$^{1,2,3\dag}$, Keji He$^{4}$, Peiyan Li$^{1,2}$ and Liang Wang$^{1,2\dag}$
\thanks{Manuscript received: May, 16, 2025; Revised November, 13, 2025; Accepted December, 12, 2025.}

\thanks{This paper was recommended for publication by Editor Aleksandra Faust upon evaluation of the Associate Editor and Reviewers' comments.}

\thanks{$\dag$ Corresponding author}
\thanks{$^{1}$New Laboratory of Pattern
Recognition (NLPR), State Key Laboratory of Multimodal Artificial Intelligence Systems (MAIS), Institute of
Automation, Chinese Academy of Sciences. $^{2}$School of Artificial Intelligence, University of Chinese Academy of Sciences. $^{3}$FiveAges. $^{4}$Shandong University.} 
\thanks{Emails: {\tt\small yixiang.chen@cripac.ia.ac.cn}, {\tt\small \{yhuang, wangliang\}@nlpr.ia.ac.cn}}
\thanks{Digital Object Identifier (DOI): see top of this page.}
}

\begin{document}
\title{
VERM: Leveraging Foundation Models to Create a \underline{V}irtual \underline{E}ye for Efficient 3D \underline{R}obotic  \underline{M}anipulation
}

\maketitle

\input{sec/0_abstract}    

\begin{IEEEkeywords}
Deep Learning for Visual Perception, Deep Learning in Grasping and Manipulation, Learning from Demonstration.
\end{IEEEkeywords}

\input{sec/1_introduction}
\input{sec/2_related_work}
\input{sec/3_VERM}
\input{sec/4_experiments}
\input{sec/5_conclusion}
\input{sec/6_acknowledgement}

\bibliographystyle{IEEEtran}
\bibliography{main}

\end{document}

%% file: sec/0_abstract.tex
\begin{abstract}
When performing 3D manipulation tasks, robots have to execute action planning based on perceptions from multiple fixed cameras. 
The multi-camera setup introduces substantial redundancy and irrelevant information, which increases computational costs and forces the model to spend extra training time extracting crucial task-relevant details.
To filter out redundant information and accurately extract task-relevant features, we propose the \textbf{VERM} (\textbf{V}irtual \textbf{E}ye for \textbf{R}obotic \textbf{M}anipulation) method, leveraging the knowledge in foundation models to imagine a virtual task-adaptive view from the constructed 3D point cloud, which efficiently captures necessary information and mitigates occlusion.
To facilitate 3D action planning and fine-grained manipulation, we further design a depth-aware module and a dynamic coarse-to-fine procedure.
Extensive experimental results on both simulation benchmark RLBench and real-world evaluations demonstrate the effectiveness of our method, surpassing previous state-of-the-art methods while achieving \textbf{1.89$\times$} speedup in training time and \textbf{1.54$\times$} speedup in inference speed. More results can be found on our project
website: \href{https://verm-ral.github.io/}{https://verm-ral.github.io/}.
\end{abstract}

%% file: sec/1_introduction.tex
\section{Introduction}
\label{introduction}
\IEEEPARstart{3}{D} robotic manipulation \cite{rt1, rt2, peract} has been widely studied due to its wide applications in industrial production and daily life. Unlike 2D pick-and-place tasks, 3D manipulation requires a comprehensive understanding of the 3D space using information from multiple fixed cameras.
Recently, new progress \cite{peract, rvt, rvt2} has been made for more accurate and efficient 3D robotic manipulation using diverse visual representations.

The specific views containing key information are crucial for robots performing 3D manipulation tasks. Robots can identify task-relevant details more clearly and mitigate potential occlusion through such views, similar to how humans choose where to interact with objects in daily life. However, in existing robotic manipulation environments, multiple fixed cameras often lead to substantial redundancy and irrelevant information in the input data, as they capture overlapping or unnecessary visual content.

Some researchers have recognized this issue and attempted to better merge information from different cameras. Some works \cite{c2f, polarnet, act3d, 3d_diffuser_actor, dp3} have projected RGB-D images from multiple cameras into a unified 3D space using either voxel or point cloud. Although these methods are able to reserve key information, the vast 3D representation still contains redundant background and irrelevant contents, making it difficult to identify the task-relevant information. Additionally, training the entire 3D representation is time-consuming, which costs about 16 days on 8 V100 GPUs \cite{peract}.
To address these issues, other works \cite{rvt, vihe, rvt2} have alternatively projected the point cloud onto predefined virtual camera planes selected by human expertise, enabling re-rendering from views containing most useful information for specific tasks. However, the selection of these camera planes relies on expert knowledge, and they might not be effective for new tasks.

Different from these methods, humans can leverage their extensive knowledge to determine the task-adaptive view for manipulation in various tasks with just a glimpse and imagination. This topic has been extensively studied in the field of cognitive science for a long time \cite{eye}, focusing on how our eyes automatically select where to look and the role of eye movements in visually guided behavior. Inspired by this, we propose the \textbf{VERM} (\textbf{V}irtual \textbf{E}ye for \textbf{R}obotic \textbf{M}anipulation) method, leveraging the knowledge in the large multimodal foundation model GPT-4o \cite{gpt4o} to guide robots in selecting the task-adaptive view from a virtual camera pose, which captures the key information for manipulation and mitigates occlusion. Specifically, VERM first predicts a virtual camera plane based on the observations from multiple fixed real cameras and textual prompts using GPT-4o and then obtains the corresponding virtual camera image from the constructed 3D point cloud. Finally, this single image is used to guide the policy for generating actions. With this \textit{predict, obtain, and guide} pipeline, the number of input tokens is significantly reduced, leading to shorter training time and faster inference speed without compromising performance.

Unlike other foundation-model-enhanced tasks such as task planning \cite{gpt4v_robotics, look_before_you_leap}, the proposed VERM overcomes two unique key challenges for 3D robotic manipulation. First, robots operate in 3D space, thus they heavily rely on depth information for spatial trajectory planning, so we design a depth-aware module to extend 2D action prediction from the single image into the third dimension. Second, robots require fine-grained manipulation to accurately complete given tasks. To address this, we propose a dynamic coarse-to-fine adjustment mechanism: when the model identifies a task-critical phase (e.g., precise alignment), it automatically triggers viewpoint zooming to focus on local regions of interest. This hierarchical refinement process allows iterative optimization of initial coarse actions only when necessary, improving manipulation success rates in high-precision tasks while maintaining efficiency.

Our contributions can be summarized as follows:

\begin{itemize}
  \item We propose a structured prompting framework that reformulates GPT-4o into a spatial reasoning agent for viewpoint selection. By encoding environment context and instructions, the model outputs grounded camera poses that capture task-relevant details and reduce occlusion, revealing the 3D spatial reasoning capability of large multimodal models.

  \item We validate the generality of our approach across different foundation models, including GPT-4o, Qwen2.5, and Claude 3.5, showing that VERM can operate as a plug-and-play spatial reasoning module without fine-tuning or architectural modification.

  \item We design a depth-aware module to extend action planning to 3D space, and a dynamic coarse-to-fine procedure to refine actions, facilitating 3D and fine-grained manipulation.
  \item The proposed VERM method demonstrates effectiveness over previous state-of-the-art methods on both simulation benchmark RLBench \cite{rlbench} and real-world evaluations, achieving \textbf{1.89$\times$} speedup in training time and \textbf{1.54$\times$} speedup in inference speed.
\end{itemize}

%% file: sec/2_related_work.tex
\section{Related Work}
\label{sec:related_work}
\noindent \textbf{Foundation Models for Robotics}~~~~
Foundation models are increasingly adopted in robotics~\cite{foundation_model_robotics, foundation_model_robot_survey}, offering prior knowledge that supports perception~\cite{clip, sam, liu2023grounding, minderer2022simple}, code generation~\cite{code_as_policies, vemprala2024chatgpt, singh2023progprompt, voxposer, rekep, vlp, vlmpc}, and task planning~\cite{rt1, rt2, gpt4v_robotics, look_before_you_leap, saycan, driess2023palm}. SayCan~\cite{saycan} decomposes human instructions into sub-tasks grounded in the physical world, while Code-as-Policies~\cite{code_as_policies} generates executable policies directly from LLMs. VoxPoser~\cite{voxposer} combines LLMs with SAM~\cite{sam} for zero-shot generalization, and foundation models like GPT-4V have enabled integrated vision-language planning~\cite{gpt4v_robotics, look_before_you_leap}.
Unlike prior works focused on high-level reasoning or skill planning, we use GPT-4o~\cite{gpt4o} to tackle viewpoint selection: generating a virtual camera pose that integrates multi-view observations and minimizes occlusion. We design a structured prompting pipeline to effectively query GPT-4o’s spatial reasoning capabilities for this purpose.

 \noindent \textbf{Vision-based Manipulation in 3D Space}~~~~The selection of an appropriate visual scene representation is crucial for vision-based manipulation tasks in the 3D space. CLIPort \cite{cliport} has used RGB-D images as input and predicted the 2D affordance map and an estimated depth value. In contrast, other works \cite{peract,c2f,act3d,3d_diffuser_actor} have alternatively employed 3D representations. Among them, C2F-ARM \cite{c2f} and Peract \cite{peract} have utilized voxel maps, while Act3D \cite{act3d} and 3D Diffuser Actor \cite{3d_diffuser_actor} have used 3D point cloud as their visual representations. These methods, though effective, can be computationally expensive. Alternatively, RVT series \cite{rvt, rvt2} and VIHE \cite{vihe} have projected RGB-D images captured from multiple cameras into a unified point cloud, which is then re-projected onto several predefined image planes. However, the selection of image planes requires human intervention and lacks flexibility for different tasks. Our work is closely related to RVT \cite{rvt} and RVT-2 \cite{rvt2}, yet we try to identify the task-adaptive view that contains all necessary information for task completion. This innovation reduces human intervention and simplifies the input to a single 2D image, thus greatly accelerating training and inference.

\noindent \textbf{High-Precision Manipulation}~~~~It is challenging to perform tasks requiring high precision. For instance, inserting pegs demands precision without any deviation. Previous research has explored various methods to achieve this. C2F-ARM \cite{c2f} has employed a coarse-to-fine approach, initially using a low-resolution voxel map to locate a coarse position, then refining the resolution around this position to enhance precision. RVT-2 \cite{rvt2} has also utilized the coarse-to-fine approach but with images as input. Building upon these foundations, our framework introduces two key innovations: (1) a single-image input architecture that reduces computational complexity and accelerates both training and inference, and (2) a \textit{dynamic} coarse-to-fine mechanism that selectively triggers fine-grained refinement only during task-critical phases. Unlike prior methods that apply refinement uniformly, our approach adapts to the task context, improving efficiency without sacrificing precision.

%% file: sec/3_VERM.tex
\section{VERM}
\label{sec:verm}

\label{sec:method}
\begin{figure*}[t]
\centering
\includegraphics[width=\linewidth]{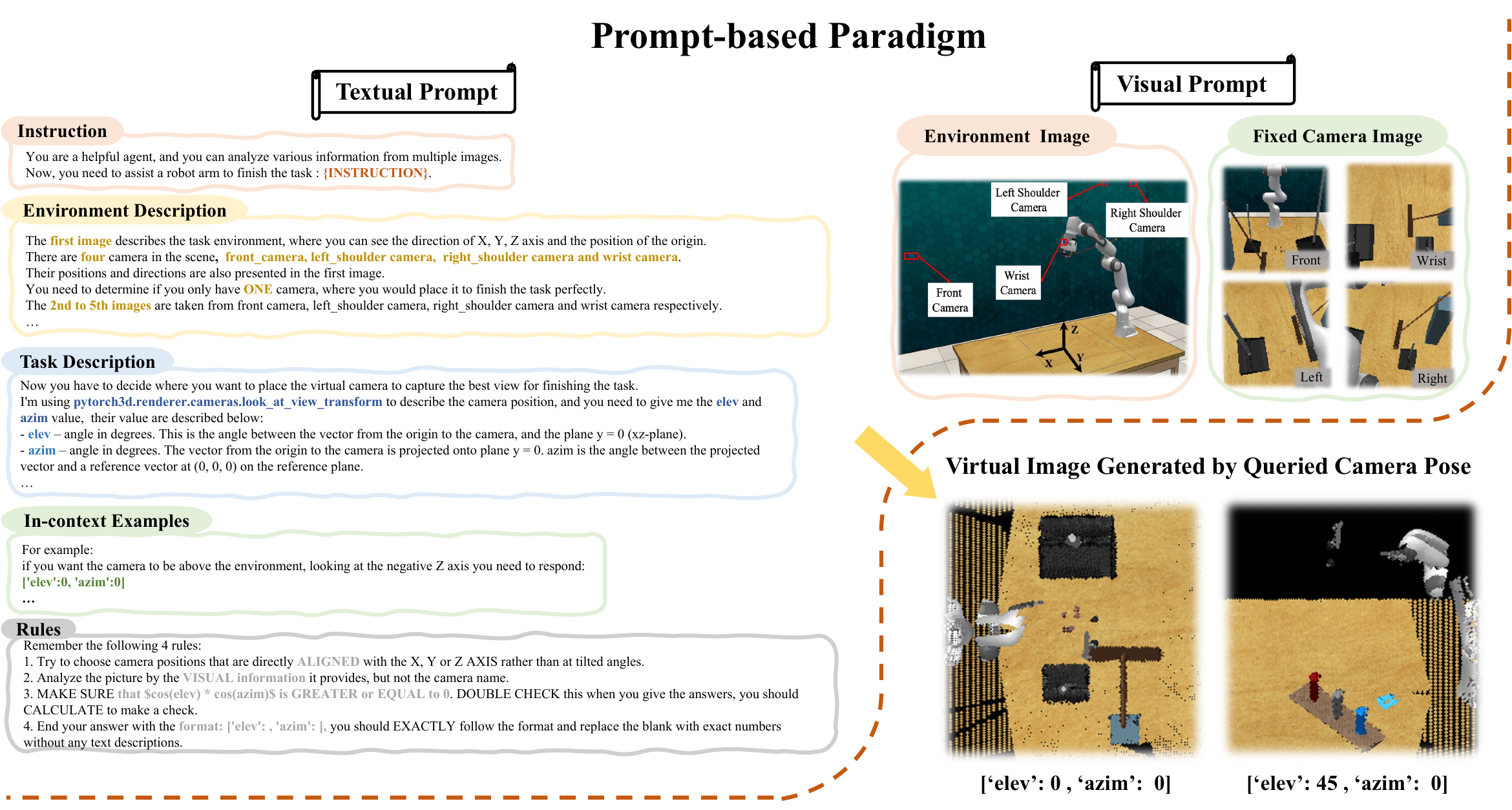}
\caption{\centering{The prompt-based paradigm for querying virtual camera poses using GPT-4o.}}
\label{fig:prompt}
\end{figure*}

\subsection{Overview}
Our objective is to perform a language-conditioned imitation learning task utilizing a dataset $D=\{\zeta_1, \zeta_2, \ldots, \zeta_n\}$ of $n$ expert demonstrations. Each demonstration $\zeta_i$ consists of sequences of RGB-D observations $\{o_{1,\ldots,m_i}^i\}$ captured from fixed scene cameras, expert action sequences $\{a_{1,\ldots,m_i}^i\}$, and a language description $l_i$, with episode length $m_i$. Our approach involves training a policy that, given natural language instructions and RGB-D observations, predicts an 8-dimensional action vector of the end-effector. This vector includes a 3-DOF position, 3-DOF rotation, an 1-DOF gripper state (open or closed), and an 1-DOF collision parameter (indicating whether the motion planner considers potential collisions with scene objects). The motion planner calculates the path between consecutive predicted actions.

\begin{figure*}[t]
\centering
\includegraphics[width=0.9\linewidth]{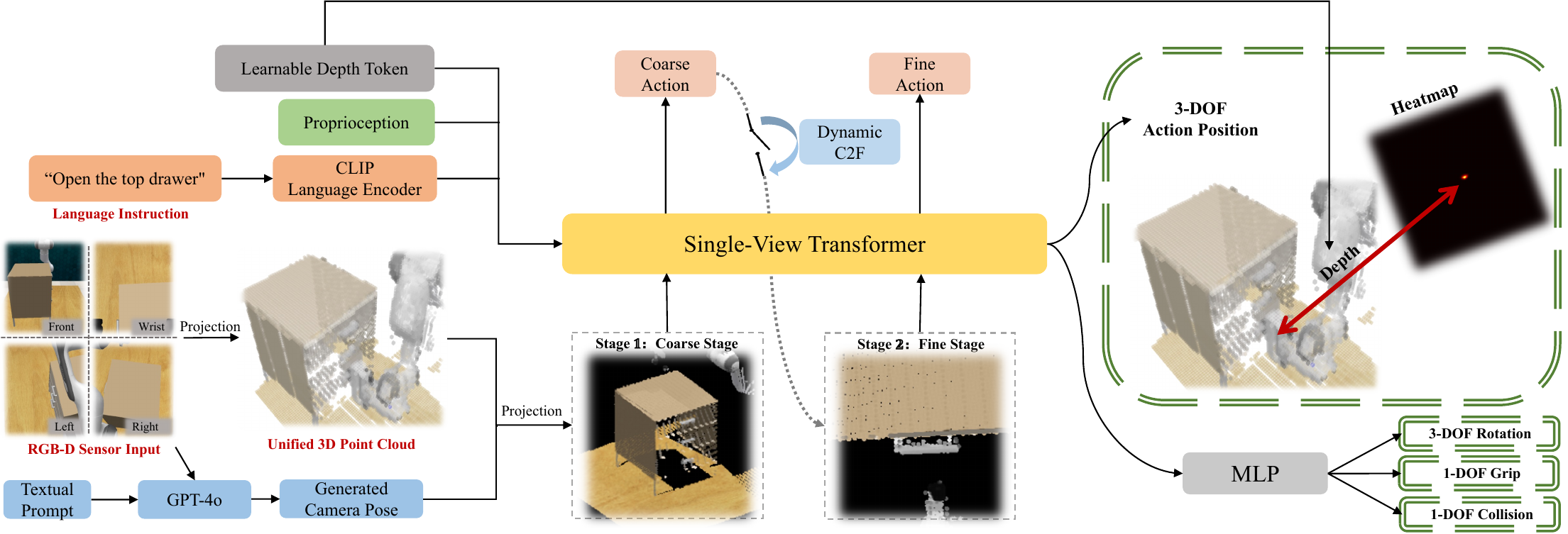}
\caption{\centering{Policy network of the proposed VERM.}}
\label{fig:network}
\end{figure*}

Our method can be summarized as follows. First, we develop a prompt-based paradigm to find the task-adaptive camera pose for completing the task using GPT-4o, as illustrated in Figure \ref{fig:prompt}. Next, given the two-dimensional nature and limited detail of the single image from the first step, we further incorporate a depth-aware module for 3D manipulation and a coarse-to-fine procedure for precise action planning, as illustrated in Figure \ref{fig:network}.

\subsection{Camera Pose Selection} \label{view_choosing}

It is non-trivial to prompt GPT-4o for querying camera poses, as it requires a comprehensive understanding of the 3D space. We develop a prompt-based paradigm for camera pose selection, which is depicted in Figure \ref{fig:prompt}. The left panel organizes a textual prompt into distinct sections, while the top-right panel presents two visual prompts, where the image on the left outlines the workspace with original fixed camera locations and axes (using the SoM (Set-of-Mark) \cite{som} prompting technique), while the right one displays original RGB images from these cameras. GPT-4o processes these prompts for camera pose selection.

We carefully structure the textual prompt into four distinct parts: environment description, task description, in-context examples, and rules.

\noindent \textbf{Environment Description} ~~~~ This section provides a basic overview of the fixed camera poses within the scene, describes the position of origin, and defines the orientations of axes in the visual prompt. Initial attempts involved specifying camera intrinsics and extrinsics, but we find that GPT-4o struggles with spatial relationships using these parameters. Instead, a visual representation of the environment using SoM could be easier for the model to interpret.

\noindent \textbf{Task Description}~~~~In this section, we define camera pose using two parameters: $\textit{elev}$ and $\textit{azim}$. The parameter $\textit{elev}$ describes the angle between the vector from the origin to the camera and the $xz-\textit{plane}$, while $\textit{azim}$ specifies the angle between the projected vector and a reference vector at $(0, 0, 0)$ on the reference plane. These angles help define the camera's direction, pointing towards the origin from a fixed distance.

\noindent \textbf{In-context Examples}~~~~We include in-context examples in the prompt to help the model better understand the spatial relationships in the environment. These examples are drawn from three cameras defined in the RVT-2 \cite{rvt} setup.

\noindent \textbf{Rules}~~~~We establish four rules to refine the output from GPT-4o to eliminate undesirable results. Firstly, we prefer camera alignments along axes to facilitate easier depth information capture. Secondly, we have observed that the model sometimes shortcuts its analysis by referring to camera labels (e.g., front\_camera) rather than the provided visual information; hence, we emphasize the importance of analyzing visual information. The third and fourth rules ensure that the camera are positioned to look downwards from above the table and constrain the output format.

To get the ideal camera pose from GPT-4o, we combine the textual prompt with five visual prompts—one for the environment description and four from original RGB-D inputs. Examples of virtual images generated by the queried camera poses are shown in the bottom-right corner of Figure 1. Although these images capture key task-relevant information, they are insufficient for 3D manipulation due to two factors: the 2D input lacks the necessary depth information for 3D action planning, and the coarse resolution cannot support fine-grained tasks like the peg insertion shown on the right. We address these limitations by incorporating a depth-aware module and a dynamic coarse-to-fine procedure in the policy network, as detailed in the following section.

\subsection{Policy Network} \label{architecture}
The architecture of the policy network is depicted in Figure \ref{fig:network}. The original RGB-D inputs are transformed into a unified 3D point cloud, which is then projected onto a virtual camera plane as specified by GPT-4o. We use this virtual image, combined with language instructions and proprioceptive (robot state and time information) inputs, to predict a coarse action. This prediction involves generating a 2D heatmap on the virtual image and estimating a corresponding depth value. Subsequently, we enable the camera to zoom in on the coarse action, allowing it to focus on the critical local content and refine the initial coarse action for higher precision.

\noindent \textbf{Dynamic Coarse-to-Fine Module}~~~~
We first transform the original RGB-D images ${o_{\textit{front}}, o_{\textit{l_shoulder}}, o_{\textit{r_shoulder}}, o_{\textit{wrist}}}$ into a unified 3D point cloud, which is then re-projected onto a virtual camera plane specified by GPT-4o’s $\textit{elev}$ and $\textit{azim}$ parameters to generate a single global image $o_{\textit{global}}$. This image merges all views to optimally capture task-relevant details. While $o_{\textit{global}}$ is sufficient for most tasks, high-precision actions benefit from finer views. In such cases, we apply a zoom-in operation by centering the point cloud on the coarse prediction and scaling it, without changing camera orientation.

However, not all stages of a task demand the same level of precision. For instance, in the \textit{stack blocks} task, free-space motions such as transporting the block can be executed with coarse predictions, whereas fine-grained accuracy is essential during grasping or placement. To handle this, we introduce a dynamic coarse-to-fine (C2F) inference module that selectively applies refinement only when needed.

Unlike prior C2F approaches that apply refinement at every step, our method uses the discrepancy between coarse and fine predictions during training to identify task-critical phases. If the predicted translation or rotation differs beyond a threshold (0.01 m or 5°), the sample is labeled as requiring refinement. A lightweight predictor is trained to classify such cases and is used at inference time to decide whether to activate the fine stage. This selective refinement strategy ensures that computational resources are focused on high-precision moments while avoiding unnecessary overhead during simpler motions, improving both accuracy and efficiency.

\noindent \textbf{Depth-Aware Module}~~~~
In this module, we incorporate learnable depth tokens to predict depth value for action planning in 3D space. The 3-DOF action position is defined on the right side of Figure \ref{fig:network}. We predict the heatmap in the same space as the observation space and estimate a depth value using depth tokens. We discretize the 3-DOF rotation into 5-degree bins for each axis to predict the rotation.

Action prediction is performed using Transformer \cite{vaswani2017attention}, which processes learnable depth tokens, language tokens, and image tokens together through an attention mechanism. 

\noindent \textbf{Keypoint Selection}~~~~
Following prior work~\cite{c2f, peract}, we reduce each trajectory to a sparse set of keypoints to avoid training on noisy intermediate steps. Keypoints are defined by two heuristics: (1) changes in end-effector state (e.g., grasp or release), and (2) near-zero velocity, typically before or after critical transitions.
% The entire task can be represented as a prediction task for the next keypoint pose. The path between two keypoints is calculated by the motion planner.

\noindent \textbf{Training}~~~~The ground truth for the heatmaps is generated from a Gaussian distribution centered on the expert action. The rotation, gripper state, and collision parameters are trained using cross-entropy loss with one-hot encoded expert actions. We also formulate depth prediction as a classification task to align it with other loss metrics. The loss is defined in Equation~\ref{formula:loss}, where $\mathcal{L}_{\textit{i}} (\textit{i} \in \{\textit{trans}, \textit{rot}, \textit{open}, \textit{collision}, \textit{depth}\})$ refers to the translation, rotation, gripper open, collision and depth loss respectively and $\mathcal{L}_{\textit{dyn\_inf}}$ refers to the loss of the dynamic coarse-to-fine indicator.
\begin{equation}
\mathcal{L} = \mathcal{L}_{\textit{trans}} + \mathcal{L}_{\textit{rot}} + \mathcal{L}_{\textit{open}} + \mathcal{L}_{\textit{collision}} + \mathcal{L}_{\textit{depth}} +
\mathcal{L}_{\textit{dyn\_inf}}\\
\label{formula:loss}
\end{equation}

%% file: sec/4_experiments.tex
\section{Experiment Results}
\label{sec:experiment}
We conduct our experiments using the simulation benchmark RLBench (Sec \ref{sec:rlbench}), as well as real-world evaluations (Sec \ref{sec:realworld}).
\subsection{RLBench}
\label{sec:rlbench}

\input{table/rlbench}

\noindent \textbf{Simulation Setup}~~~~
Our experimental setup on RLBench \cite{rlbench} is designed in line with PerAct \cite{peract}, which uses CoppeliaSim \cite{rohmer2013v} to simulate a variety of tasks from RLBench. We employ a Franka Panda robot equipped with a parallel gripper to perform tasks introduced by PerAct, which include diverse activities such as picking and placing, tool manipulation, drawer opening, and high-accuracy peg insertions. For each task, variations are created based on associated language descriptions to enhance task complexity. Visual observations are captured from four noiseless RGB-D cameras placed at strategic locations (front, left shoulder, right shoulder, and wrist), each providing $128\times128$ resolution images. The path between predicted action and present action is calculated using a sampling-based motion planner \cite{karaman2011sampling}, as used in \cite{c2f,peract,act3d,rvt,vihe}, which facilitates the generation of joint space actions needed to complete the tasks effectively.

\noindent \textbf{Compared Methods}~~~~
We compare our method against eight state-of-the-art methods and all methods use the same RGB-D inputs from four fixed cameras for fair comparison: 
\begin{enumerate}
\item [(1)]C2F-ARM-BC \cite{c2f}, which has converted RGB-D images into multi-resolution voxel representations and predicted key-frame actions through a coarse-to-fine strategy; 

\item [(2)]PerAct \cite{peract}, which has voxelized RGB-D images and employed a Perceiver transformer for action prediction; 

\item [(3)]HiveFormer \cite{hiveformer}, which has directly used images captured from original cameras; 

\item [(4)]PolarNet \cite{polarnet}, which has used dense point representation for action prediction;

\item [(5)]RVT \cite{rvt}, which has generated five global orthographic views by human expertise and utilizes a multi-view transformer for predicting actions; 

\item [(6)]Act3D \cite{act3d}, which has leveraged pre-trained image features and applied relative cross-attention mechanisms on point cloud for action detection;

\item [(7)]3D Diffuser Actor (3DDA) \cite{3d_diffuser_actor}, which has used diffusion policy to predict actions on 3D point cloud;

\item [(8)]RVT-2 \cite{rvt2}, which has been built on RVT \cite{rvt} with more efficient implementation and coarse-to-fine strategy.
\end{enumerate}

\noindent \textbf{Implementation Details}~~~~
We use the ResNet50 variant of CLIP \cite{clip} to encode language instructions. Images are rendered from point clouds via PyTorch3D orthographic cameras \cite{rvt} at $224\times224$ resolution. The Transformer takes $256$ image tokens (patch size $14\times14$), $77$ language tokens, and $36$ learnable depth tokens, with 8 layers. Camera view selection is performed once per task using the OpenAI \texttt{gpt-4o} API.

% We apply data augmentation on point clouds using perturbations in the range $[\pm 0.125m, \pm 0.125m, \pm 0.125m]$ for translation and $\pm 45^{\circ}$ for rotations around the z-axis.

\noindent \textbf{Training and Evaluation Details}~~~~
Each RLBench task contains 100 expert demonstrations, collected using four fixed RGB-D cameras per scene. Each trajectory consists of 50–150 timesteps, from which we extract 2–12 keypoints.
Training is conducted on 8 NVIDIA V100 (16GB) GPUs using the LAMB optimizer \cite{you2019large}, with batch size 640 ($80 \times 8$) and learning rate $2.4 \times 10^{-3}$. Evaluation follows PerAct \cite{cliport}, using 25 trajectories per task. A trial is marked successful if the robot meets predefined conditions; failure occurs if the planner yields invalid actions or exceeds time limits. To account for randomness in sampling-based planning, we repeat each trial five times and report both mean and variance of success rates.
\begin{figure*}[ht]
\centering
\includegraphics[width=\linewidth]{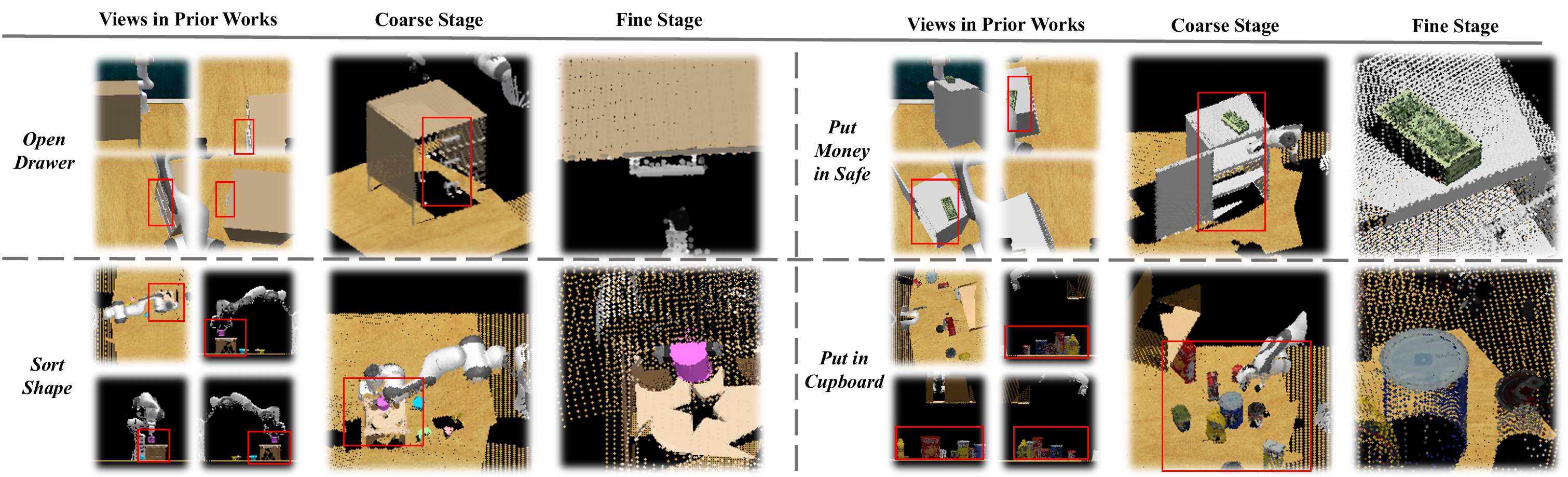}
\caption{\centering{Visualization of action prediction of VERM in RLBench.}}
\label{fig:visualization_results}
\end{figure*}

\begin{figure}[h]
\centering
\begin{minipage}[c]{0.48\linewidth}
    \centering
    \includegraphics[width=\textwidth]{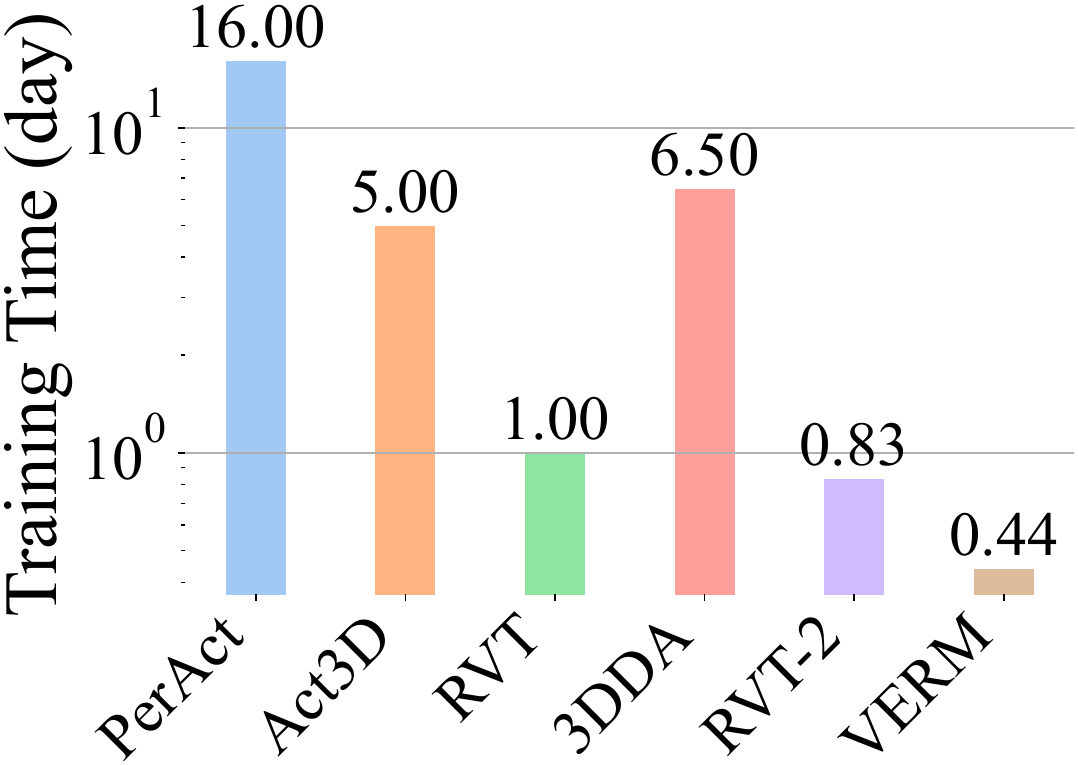}
    % \caption{Training Time (Days) in Log Scale.}
    % \label{fig:training_time}
\end{minipage}
\begin{minipage}[c]{0.48\linewidth}
    \centering
    \includegraphics[width=\textwidth]{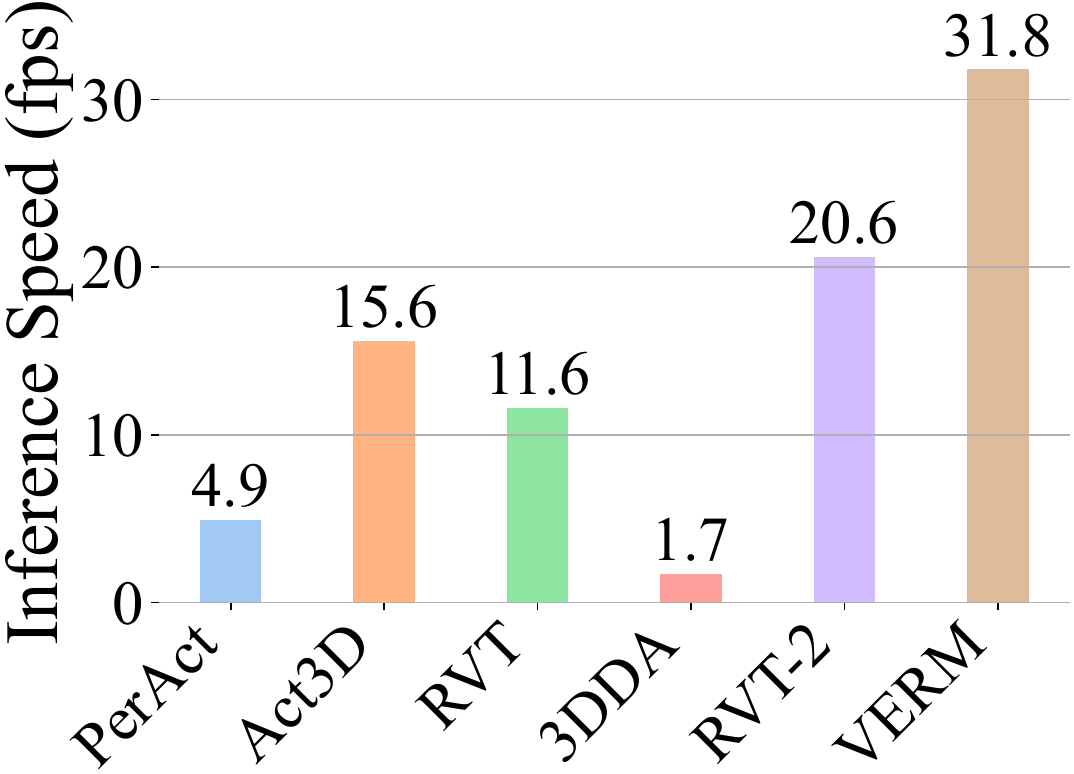}
    % \caption{Inference Speed (fps).}
    % \label{fig:inference_speed}
\end{minipage} 
\caption{\textit{Left}: Training time (day) in log scale. \textit{Right}: Inference speed (fps).}
\label{fig:time_camparison}
\end{figure}

\noindent \textbf{Results}~~~~
The training time and inference speed of VERM, along with those of previous works\footnote{PerAct, RVT, RVT-2, and VERM use 8 16GB V100 GPUs, while Act3D uses 8 32GB V100 GPUs and 3DDA uses 7 40GB A100 GPUs. Among these methods, VERM aligns with the least demanding GPU setup.}, are presented in Figure \ref{fig:time_camparison}. VERM achieves \textbf{1.89$\times$} speedup in training time and \textbf{1.54$\times$} speedup in inference speed than the previous state-of-the-art method RVT-2 using the same GPU setup, which demonstrates that single-image input not only enables efficient learning but also opens up new possibilities for real-time control.

In addition to the reduced training time and faster inference speed, VERM does not compromise task success rate, as shown in Table \ref{tab:rlbench}. VERM surpasses RVT-2 by 1.4\% in average task success rate and performs best in 11 out of 17 tasks. This indicates that the single image, informed by GPT-4o, is sufficient for acquiring all task-relevant content. VERM  alleviates the need to pre-define multiple virtual camera planes in the RVT series method (Note that the 3DDA method adopts an original image size of $256$, whereas VERM and the RVT series use $128$).

\noindent \textbf{Qualitative Analysis}~~~~
Figure~\ref{fig:visualization_results} shows that VERM-generated views effectively integrate information from multiple fixed cameras while avoiding occlusion. In \textit{open drawer} and \textit{put money}, the selected view reveals the full object (e.g., entire handle) that is partially missing in original views. In \textit{sort shape}, a slight rotation exposes occluded holes. These results demonstrate VERM’s ability to generate concise, task-relevant views that support accurate action prediction.

\input{table/ablation}

\begin{figure}[t]
  \centering
  \includegraphics[width=\linewidth]{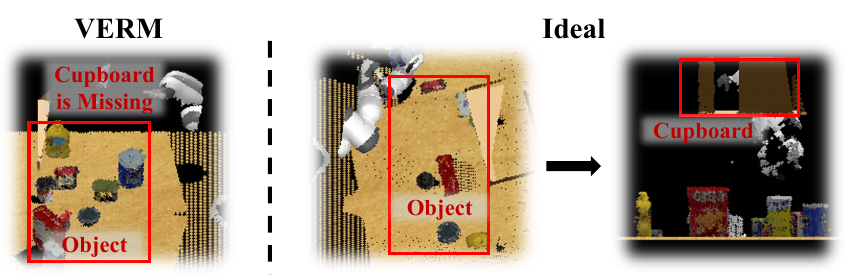}
  \caption{\centering{Example failure cases.}}
  \label{fig:failure_cases}
\end{figure}

\begin{figure*}[t]
\centering
\includegraphics[width=\linewidth]{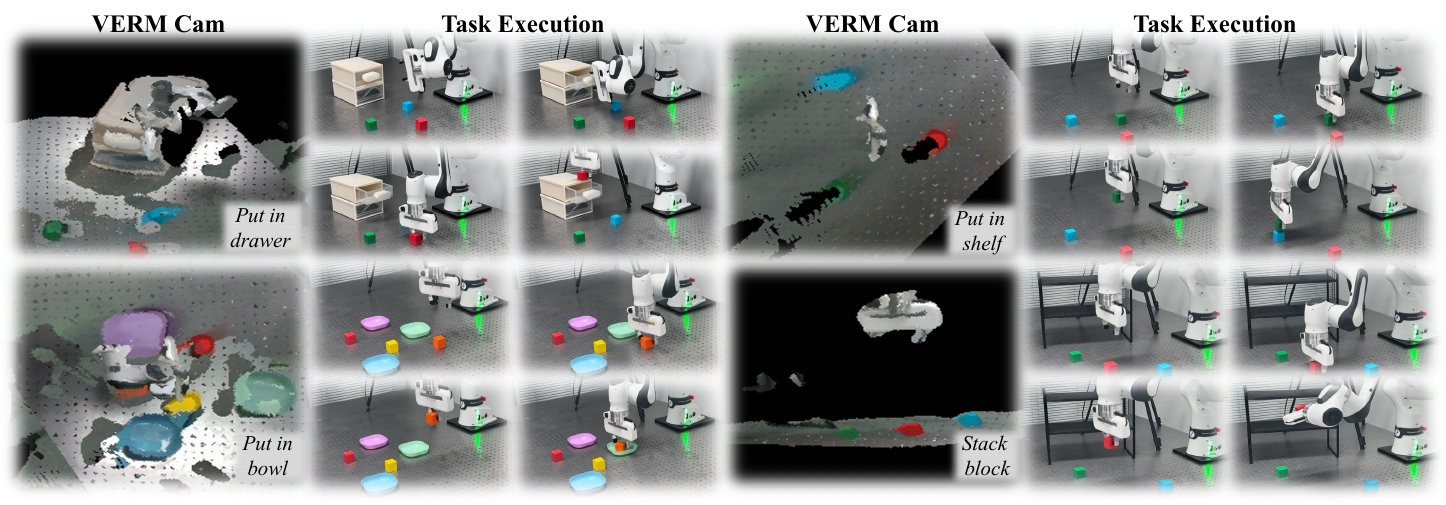}
\caption{\centering{Visualization of action prediction of VERM in real-world.}}
\label{fig:realworld_vis}
\end{figure*}

\noindent \textbf{Ablation Study}~~~~
The ablation results in Table~\ref{tab:ablation} show that model \#1 represents the full VERM design. Models \#2–\#4 test single-camera inputs from the RVT-2 setup, where the GPT-4o–generated global view outperforms all predefined cameras by adapting to task variations. The top camera performs best among the fixed ones but remains limited under occlusion or multi-object orientations. Removing the coarse-to-fine process (\#5) reduces precision, while the rendering settings in \#6 and \#7 confirm that low resolution and missing zoom-in both degrade performance.
Model \#8 removes the axis-alignment constraint, leading to a performance drop, indicating that alignment improves depth prediction while still allowing necessary rotations to reduce occlusion.

\noindent \textbf{Cross-Model Generalization}~~~~
To evaluate the generalizability of VERM across different foundation models, we additionally tested our method using Qwen2.5 (open-source) \cite{qwen} and Claude 3.5 Sonnet \cite{claude35}. Since these models tend to generate suboptimal poses on the first attempt, we apply an iterative refinement strategy with self-verification feedback to adjust camera parameters. As shown in Table~\ref{tab:abla_vlm}, both models achieve comparable task success rates (80.3\% and 81.2\%, respectively) to GPT-4o (83.6\%), despite moderate angular deviations. These results confirm that VERM is compatible with alternative foundation models while maintaining robust performance.
\input{table/abla_VLM}

\noindent \textbf{Failure Cases and Model Hallucinations}~~~~
Figure~\ref{fig:failure_cases} shows typical failure cases. In tasks like \textit{put in cupboard}, the object and target container are visible from different angles, making it difficult to capture all relevant details in a single view. As our method queries the foundation model only once at the task's start, it may miss critical cues needed in later stages.
We address this with a dynamic re-querying strategy that updates the viewpoint mid-execution, raising success of \textit{put in cupboard} task from 55.2\% to 66.4\%, although with higher computational cost.

We also observe hallucinations from GPT-4o, such as invalid viewpoints (e.g., under the table) or overreliance on camera labels instead of visual cues. To reduce this, we enforce prompt constraints: (i) reasoning must be based on visual input only, and (ii) generated poses must satisfy geometric validity ($z > 0$). A self-verification loop further filters out unsatisfactory views by checking the rendered result against criteria. These measures improve robustness and support generalization to other models like Qwen and Claude, which tend to hallucinate more frequently.

\subsection{Real-world Evaluation}
\label{sec:realworld}
\noindent \textbf{Real-world Setup}~~~~
We evaluate our VERM methods by comparing them with RVT \cite{rvt} and RVT-2\cite{rvt2} on a real-world manipulation setup, similar to those used in previous studies. This setup includes a statically mounted Franka Research 3 arm and 2 stationary Intel Realsense D435i RGB-D cameras for third-person viewing. We calibrate the robot camera extrinsic and convert the two perceived point clouds into the unified robot's base frame before feeding them into the VERM network. For a given target gripper action, we utilize Deoxys \cite{deoxys} to guide the robot to the target through trajectory generation and feedback control.

\noindent \textbf{Task Definition and Data Collection}~~~~
We evaluate VERM on eight real-world manipulation tasks: \textit{stack blocks}, \textit{press sanitizer}, \textit{place block in bowl}, \textit{place object in drawer}, \textit{place object in shelf}, \textit{flip cup}, \textit{close drawer}, and \textit{open cabinet}. Each task includes 1–7 object variants defined by language instructions. Each task contains 100 demonstrations, augmented via 3D point cloud transformations to simulate diverse spatial layouts. Evaluation is conducted over 10 trials per configuration using unseen object placements. 
% Following the simulation environment RLBench, we define task-specific keypoints and program the Franka Panda arm to sequentially reach these waypoints. 
% At each keypoint position, we record RGB-D observations of two side cameras and actions.

\noindent \textbf{Results}~~~~
Table~\ref{tab:real_world} reports the success rates under two settings: using 15 and 100 demonstrations per task. VERM achieves strong performance with just 15 trajectories, already outperforming RVT and RVT-2 in most tasks while significantly reducing both training time and inference latency. When trained with 100 demonstrations, performance further improves slightly, suggesting that our method is highly data-efficient and generalizes well with limited supervision.
\input{table/realworld}

Figure~\ref{fig:realworld_vis} shows action predictions in the real world. Despite using low-cost RGB-D sensors and encountering lighting or point cloud noise, the virtual views generated by VERM still preserve key task-relevant details, enabling reliable execution.

%% file: table/rlbench.tex
\begin{table*}[t]
    \caption{Multi-task performance on RLBench.}
    \centering
    \begin{adjustbox}{width=\textwidth}
    \setlength{\tabcolsep}{0.5mm}
    \begin{tabular}{@{}lccccccccc@{}}
    \toprule
    & Avg. & Stack & Open & Slide & Sweep to & Meat off & Turn & Put in & Close \\
    & Success $\uparrow$ & Blocks & Drawer & Block & Dustpan & Grill & Tap & Drawer & Jar \\
    \midrule
    C2F-ARM-BC~\cite{c2f} & 16.9 & 0 & 20.0 & 16.0 & 0.0 & 20.0 & 68.0 & 4.0 & 24.0\\
    PerAct~\cite{peract} & 52.3 & 26.4$_{\pm 3.2}$ & 88.0$_{\pm 5.7}$ & 74.0$_{\pm 13.0}$ & 52.0$_{\pm 0.0}$ & 70.4$_{\pm 2.0}$ & 88.0$_{\pm 4.4}$ & 51.2$_{\pm 4.7}$ & 55.2$_{\pm 4.7}$\\
    HiveFormer~\cite{hiveformer} & 48.0 & 8.0  & 52.0 & 64.0 & 28.0 & \textbf{100.0} & 80.0 & 68.0 & 52.0\\
    PolarNet~\cite{polarnet} & 48.7 & 4.0 & 84.0 & 56.0 & 52.0 & \textbf{100.0} & 80.0 & 32.0 & 36.0\\
    RVT~\cite{rvt} & 65.0 & 28.8$_{\pm 3.9}$ & 71.2$_{\pm 6.9}$ & 81.6$_{\pm 5.4}$ & 72.0$_{\pm 0.0}$ & 88.0$_{\pm 2.5}$ & 93.6$_{\pm 4.1}$ & 88.0$_{\pm 5.7}$ & 52.0$_{\pm 2.5}$\\
    Act3D~\cite{act3d} & 68.4 & 12.0 & 93.0 & 93.0 & 92.0 & 94.0  & 94.0 & 90.0  & 93.0 \\
    3DDA~\cite{3d_diffuser_actor} & 83.5  & 68.3$_{\pm 3.3}$ & \textbf{89.6}$_{\pm 4.1}$ & 97.6$_{\pm 3.2}$ & 84.0$_{\pm 4.4}$ & 96.8$_{\pm 1.6}$ & \textbf{99.2}$_{\pm 1.6}$ & 96.0$_{\pm 3.6}$ & 96.0$_{\pm 2.5}$\\
    RVT-2~\cite{rvt2} & 82.2 & 80.0$_{\pm 2.86}$ & 74.0$_{\pm 11.8}$ & 92.0$_{\pm 2.8}$& \textbf{100.0}$_{\pm 0.0}$ & 99.0$_{\pm 1.7}$& 99.0$_{\pm 1.7}$ & 96.0$_{\pm 0.0}$ & \textbf{100.0}$_{\pm 0.0}$ \\
    VERM(ours)~ & \textbf{83.6} & \textbf{80.8}$_{\pm 4.4}$ & \textbf{89.6}$_{\pm 3.8}$ & \textbf{99.2}$_{\pm 2.6}$  & 95.2$_{\pm 1.8}$ & \textbf{100.0}$_{\pm 0.0}$  & 98.4$_{\pm 2.2}$  & \textbf{100.0}$_{\pm 0.0}$  & 96.8$_{\pm 3.3}$ \\
    \midrule\midrule
     & Screw & Put in & Place & Put in & Sort & Push & Insert & Place & Drag \\
    & Bulb & Safe & Wine & Cupboard & Shape & Buttons & Peg & Cups & Stick\\
    \midrule
    C2F-ARM-BC~\cite{c2f} & 8.0 & 12.0 & 8.0 & 0.0 & 8.0 & 72.0 & 4.0 & 0.0 & 24.0 \\
    PerAct~\cite{peract}  & 17.6$_{\pm 2.0}$ & 86.0$_{\pm 3.6}$ & 44.8$_{\pm 7.8}$ & 28.0$_{\pm 4.4}$ & 16.8$_{\pm 4.7}$ & 92.8$_{\pm 3.0}$ & 5.6$_{\pm 4.1}$ & 2.4$_{\pm 2.2}$ & 89.6$_{\pm 4.1}$\\
    HiveFormer~\cite{hiveformer} & 8.0  & 76.0 & 80.0 & 32.0 & 8.0 & 84.0 & 0.0 & 0.0 & 76.0\\
    PolarNet~\cite{polarnet}  & 44.0 & 84.0 & 40.0 & 12.0 & 12.0 & 96.0 & 4.0  & 0.0 & 92.0\\
    RVT~\cite{rvt}  & 48.0$_{\pm 5.7}$ & 91.2$_{\pm 3.0}$ & 91.0$_{\pm 5.2}$ & 49.6$_{\pm 3.2}$ & 36.0$_{\pm 2.5}$ & 100.0$_{\pm 0.0}$ & 11.2$_{\pm 3.0}$ & 4.0$_{\pm 2.5}$ & 99.2$_{\pm 1.6}$\\
    Act3D~\cite{act3d}  & 47.0 & 95.0 & 80.0 & 51.0 & 8.0 & 99.0 & 27.0  & 3.0 & 92.0 \\
    3DDA~\cite{3d_diffuser_actor} &  $82.4_{\pm 2.0}$ & $97.6_{\pm 2.0}$ & $93.6_{\pm 4.8}$ & $\textbf{85.6}_{\pm 4.1}$ & $44.0_{\pm 4.4}$ & $98.4_{\pm 2.0}$ & $\textbf{65.6}_{\pm 4.1}$ & $24.0_{\pm 7.6}$ & $\textbf{100.0}_{\pm 0.0}$\\
    RVT-2~\cite{rvt2}  & 88.0$_{\pm 4.9}$ & 96.0$_{\pm 2.8}$ & 95.0$_{\pm 3.3}$&  66.0$_{\pm 4.5}$ & 35.0$_{\pm 7.1}$ & \textbf{100.0}$_{\pm 0.0}$ & 40.0$_{\pm 0.0}$ & 38.0$_{\pm 4.5}$ & 99.0$_{\pm 1.7}$\\
    VERM(ours)~  & \textbf{93.6}$_{\pm 3.6}$ & \textbf{100.0}$_{\pm 0.0}$ & \textbf{96.0}$_{\pm 2.8}$  & 55.2$_{\pm 6.6}$ & \textbf{47.2}$_{\pm 5.9}$  & \textbf{100.0}$_{\pm 0.0}$  & 30.4$_{\pm 4.6}$  & \textbf{40.0}$_{\pm 2.8}$ & 99.2$_{\pm 1.8}$\\
    \bottomrule
    \end{tabular}
    \end{adjustbox}
\label{tab:rlbench}
\end{table*}

%% file: table/ablation.tex
\begin{table}[t]
\caption{Ablation study.}
\small
\centering
\begin{adjustbox}{width=0.9\linewidth}
\begin{tabular}{ccccccccccc}
\toprule
\multicolumn{1}{c}{\multirow{3}{*}{\#}} & \multicolumn{5}{c}{Core Design} & \multicolumn{4}{c}{Rendering Parameters} & \multicolumn{1}{c}{\multirow{3}{*}{Avg. Succ.}}\\
\cmidrule(r){2-6} \cmidrule(lr){7-10}
& \multicolumn{4}{c}{Cam. Pose} & \multicolumn{1}{c}{\multirow{2}{*}{C2F}} & \multicolumn{2}{c}{Img. Size} & Zoom & Align & \multicolumn{1}{c}{}\\
\cmidrule(r){2-5} \cmidrule(lr){7-8}
& GPT-4o & Front  & Top & Right &  & 112 & 224  & In & Axis \\
\midrule
1& \Checkmark  &  &  &  & \Checkmark &  & \Checkmark &\Checkmark &\Checkmark &  \textbf{83.6}\\
2& & \Checkmark &  &  & \Checkmark &  & \Checkmark &\Checkmark &\Checkmark & 58.3 \\
3& &  & \Checkmark &  & \Checkmark &  & \Checkmark &\Checkmark &\Checkmark &  70.9 \\
4& &  & & \Checkmark & \Checkmark &  & \Checkmark &\Checkmark &\Checkmark & 55.1\\
5& \Checkmark  &  &  &  &  &  & \Checkmark &\Checkmark &\Checkmark & 56.7\\
6& \Checkmark  &  &  &  & \Checkmark & \Checkmark &  &\Checkmark &\Checkmark & 71.3\\
7& \Checkmark  &  &  &  & \Checkmark & \Checkmark & \Checkmark & &\Checkmark & 59.2\\
8 & \Checkmark &  &  &  & \Checkmark &  & \Checkmark & \Checkmark &  & 66.4\\
\bottomrule
\end{tabular}
\end{adjustbox}
\label{tab:ablation}
\end{table}\

%% file: table/abla_VLM.tex
\begin{table}[h]
\caption{Evaluation of Camera Pose Generation Capabilities of Different Foundation Models.}
    \centering
    \begin{adjustbox}{width=0.8\linewidth}
    \begin{tabular}{@{}lcc@{}}
        \toprule
         Model & Deviation from GPT-4o & Success Rate \\
         \midrule
         GPT-4o \cite{gpt4o}& 0° & \textbf{83.6\%} \\
         Qwen2.5 \cite{qwen} & 11° & 80.3\% \\
         Claude 3.5 Sonnet \cite{claude35} & 8° & 81.2\% \\
        \bottomrule
    \end{tabular}
    \end{adjustbox}
\label{tab:abla_vlm}
\end{table}

%% file: table/realworld.tex
\begin{table}[htb]
\caption{Multi-task performance on real-world evaluation.}
\centering
\begin{adjustbox}{width=\linewidth}
\begin{tabular}{@{}lccccc@{}}
\toprule
\multirow{3}{*}{Task} & \# of & \multicolumn{4}{c}{Models} \\
\cmidrule{3-6}
 & \multirow{2}{*}{vari.} & RVT & RVT-2 & VERM & VERM \\
 & & (15 Traj) & (15 Traj) & (15 Traj) & (100 Traj) \\
\midrule
Stack blocks & 3 & 70\% & \textbf{80\%} & \textbf{80\%} & \textbf{80\%} \\
Press sanitizer & 1 & 70\% & 80\% & \textbf{90\%} & 80\% \\
Put block in bowl & 5 & 40\% & 60\% & 70\% & \textbf{80\%} \\
Put object in drawer & 7 & 30\% & 50\% & \textbf{70\%} & \textbf{70\%} \\
Put object in shelf & 7 & 50\% & 70\% & \textbf{80\%} & \textbf{80\%} \\
Flip cup & 1 & 20\% & 60\% & \textbf{80\%} & 70\% \\
Close drawer & 1 & 50\% & 70\% & 90\% & \textbf{100\%} \\
Open cabinet & 1 & 40\% & 70\% & 70\% & \textbf{80\%} \\
\midrule
All tasks & 26 & 46.25\% & 67.50\% & 78.75\% & \textbf{80.00\%} \\
\bottomrule
\end{tabular}
\end{adjustbox}
\label{tab:real_world}
\end{table}

%% file: sec/5_conclusion.tex
\section{Conclusion}
\label{sec:conclusion}
We present VERM, a framework that leverages GPT-4o to generate a virtual eye for robotic manipulation. By fusing multi-camera inputs into a task-adaptive view, VERM reduces perception to a single image while maintaining high success rates. A dynamic coarse-to-fine module further refines actions, enabling efficient and accurate manipulation.

\noindent \textbf{Limitations and Future Work}~~~~
Our approach assumes fixed multi-camera setups and queries the viewpoint only at the beginning. Future work may explore dynamic view selection over time for better adaptability, as well as incorporating history observations to support long-horizon planning. 
We also plan to evaluate VERM on contact-rich manipulation tasks using datasets like REASSEMBLE \cite{resemble}, to further assess generalization under physical interaction constraints.

%% file: sec/6_acknowledgement.tex
\section{ACKNOWLEDGEMENT}
This work was jointly supported by National Natural Science Foundation of China (62322607, 62236010 and 62276261), Beijing Natural Science Foundation (L252033), Taishan Scholar Foundation of Shandong Province (tsqn202507043), Natural Science Foundation of Shandong Province (ZR2025QC1566) and FiveAges Grant.